\documentclass[10pt,twocolumn,letterpaper]{article}

\usepackage{iccv}
\usepackage{times}
\usepackage{epsfig}
\usepackage{graphicx}
\usepackage{amsmath}
\usepackage{amssymb}
\usepackage{setspace}
\usepackage[bb=px]{mathalfa} 
\usepackage{adjustbox}
\usepackage{bigints}
\graphicspath{{./images/}}

\usepackage{graphicx}
\usepackage{amsmath}
\usepackage{amssymb}
\usepackage{booktabs}
\usepackage[utf8]{inputenc} 
\usepackage[T1]{fontenc}    
\usepackage{url}            
\usepackage{booktabs}       
\usepackage{amsfonts}       
\usepackage{nicefrac}       
\usepackage{microtype}      
\usepackage{xcolor}         
\usepackage[linesnumbered,ruled,vlined]{algorithm2e}

\usepackage{array}
\usepackage{amssymb}
\usepackage{latexsym}
\usepackage{amsmath,amssymb}
\usepackage{graphicx}
\usepackage{multirow}
\usepackage{adjustbox}
\usepackage{amsfonts}   
\usepackage{makecell}
\usepackage[nottoc]{tocbibind}\usepackage{url}
\usepackage{wrapfig}

\SetCommentSty{mycommfont}
\newlength{\commentWidth}
\usepackage{caption}
\usepackage{listings}
\newcommand{\twopartdef}[4]
{\setstretch{2.25}
	\left\{
		\begin{array}{ll}
			#1 & \mbox{if } #2 \\ 
			#3 & \mbox{if } #4
		\end{array}
	\right.
}

\usepackage[breaklinks=true,bookmarks=false]{hyperref}
\hypersetup{
    colorlinks=true,
    linkcolor=red,
    filecolor=magenta,      
    urlcolor=magenta,
}

\iccvfinalcopy 

\def\iccvPaperID{XXXX} 
\def\httilde{\mbox{\tt\raisebox{-.5ex}{\symbol{126}}}}

\begin{document}

\title{DITTO-NeRF: Diffusion-based Iterative Text To Omni-directional 3D Model}

\author{Hoigi Seo$^{1*}$ \qquad Hayeon Kim$^{1*}$ \qquad Gwanghyun Kim$^{1*}$ \qquad \stepcounter{footnote} Se Young Chun$^{1,2}$\thanks{} \\
$^1$Dept. of Electrical and Computer Engineering, $^2$INMC \&  IPAI \\
Seoul National University, Republic of Korea \\
{\tt\small \{seohoiki3215, khy5630, gwang.kim, sychun\}@snu.ac.kr}
}

\begin{figure*}
\twocolumn[{
\maketitle

\begin{center}
\vspace*{-2.em}
\includegraphics[width=2\columnwidth]{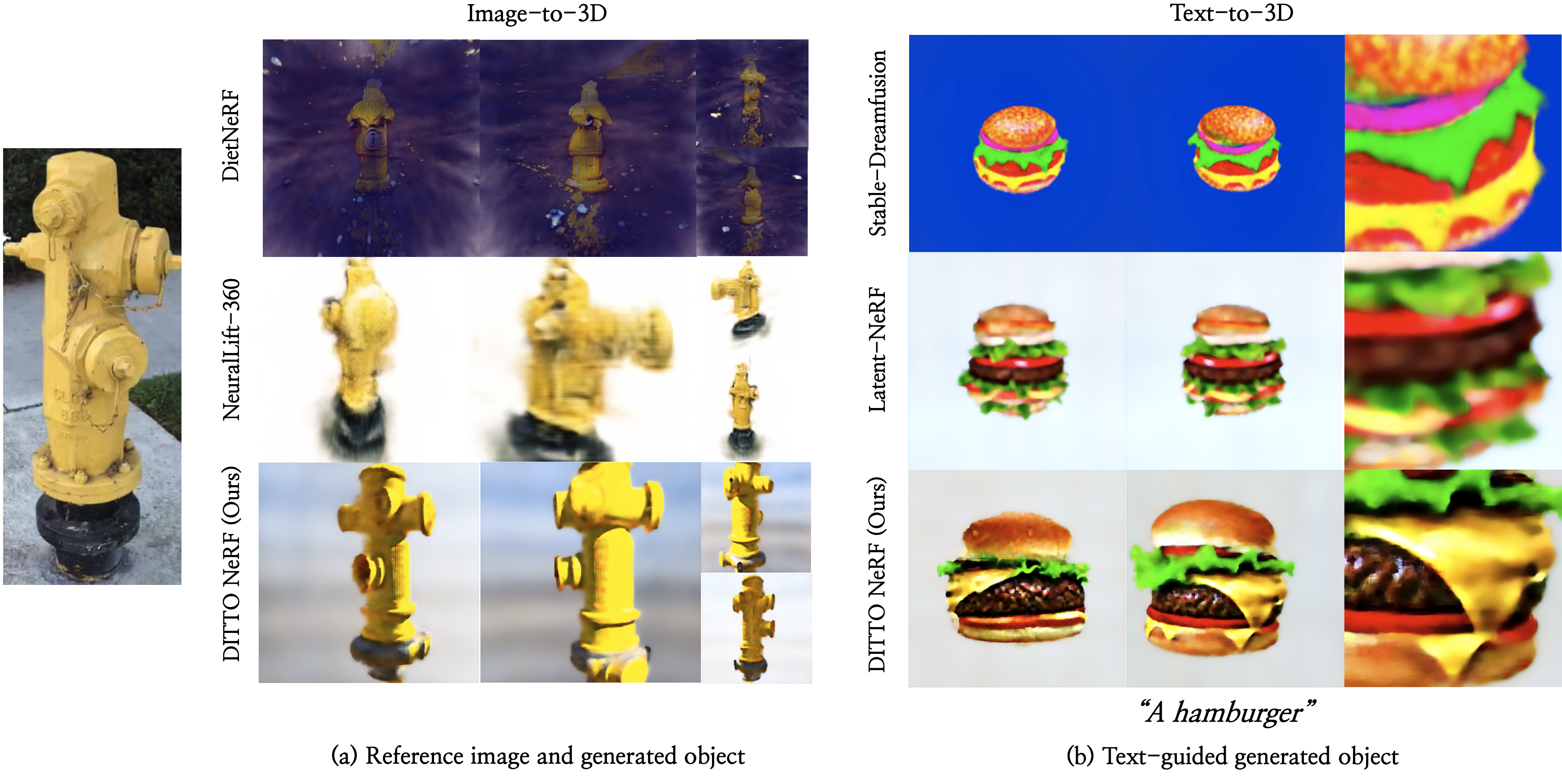}
\vspace*{-1.em}
\newline
\caption{3D NeRF models generated from (a) a reference image along with the text ``\textit{a yellow fire hydrant}'' and (b) the text ``\textit{a hamburger}'' using our DITTO-NeRF and prior arts (DietNeRF~\cite{jain2021putting}, NeuralLift-360~\cite{Xu_2022_neuralLift}, Latent-NeRF~\cite{metzer2022latent}, Stable-DreamFusion~\cite{stable-dreamfusion}). In (b), photos are zoomed in the rightmost column for comparisons. Our DITTO-NeRF yielded exquisite images in all views from text and/or image. See the supplementary videos at
\href{https://janeyeon.github.io/ditto-nerf}{{\texttt{janeyeon.github.io/ditto-nerf}}}.}

\label{fig:short}
\label{fig:front_results}
\end{center}
}]
\end{figure*}

\def\thefootnote{*}\footnotetext{Equal contribution. $^\dagger$Corresponding author.}\def\thefootnote{\arabic{footnote}}

\textheight 9in
\begin{abstract}
The increasing demand for high-quality 3D content creation has motivated the development of automated methods for creating 3D object models from a single image and/or from a text prompt. However, the reconstructed 3D objects using state-of-the-art image-to-3D methods still exhibit low correspondence to the given image and low multi-view consistency. Recent state-of-the-art text-to-3D methods are also limited, yielding 3D samples with low diversity per prompt with long synthesis time. To address these challenges, we propose DITTO-NeRF, a novel pipeline to generate a high-quality 3D NeRF model from a text prompt or a single image. Our DITTO-NeRF consists of constructing high-quality partial 3D object for limited in-boundary (IB) angles using the given or text-generated 2D image from the frontal view and then iteratively reconstructing the remaining 3D NeRF using inpainting latent diffusion model. We propose progressive 3D object reconstruction schemes in terms of scales (low to high resolution), angles (IB angles initially to outer-boundary (OB) later), and masks (object to background boundary) in our DITTO-NeRF so that high-quality information on IB can be propagated into OB. Our DITTO-NeRF outperforms state-of-the-art methods in terms of fidelity and diversity qualitatively and quantitatively with much faster training times than prior arts on image/text-to-3D such as DreamFusion, and NeuralLift-360.
\end{abstract}

\section{Introduction}

Recent advancements in virtual reality and augmented reality have led to a rapid increase in demand for 3D content.
Nevertheless, traditionally the creation of high-quality 3D objects has been a time-consuming and costly process that requires human experts.   
The challenge of the high cost of making 3D objects has motivated to development of methods for synthesizing diverse 3D objects from simplified source inputs. The methods of generating 3D objects from a single image (image-to-3D)~\cite{durou2008numerical, fan2017point, favaro2005geometric, johnston2017scaling, li2018megadepth, saito2019pifu, wang2018pixel2mesh, wu2017marrnet, Xu_2022_SinNeRF, kangle2021dsnerf, yu2021pixelnerf, Xu_2022_neuralLift} have been developed. 
The creation of a 3D object from a single image is a challenging task that is hindered by the insufficiency of available information. Recent image-to-3D studies such as DietNeRF~\cite{jain2021putting} and NeuralLift-360~\cite{Xu_2022_neuralLift} have shown the impressive results leveraging the prior knowledge of pre-trained  CLIP (Contrastive Language-Image Pre-training)~\cite{radford2021learning} models or text-to-image diffusion models. Other approaches are the methods of generating 3D objects from a text prompt (text-to-3D)~\cite{poole2022dreamfusion, metzer2022latent, lin2022magic3d, wang2022clip} by leveraging text-to-image model to create 3D representation.

 Recent image-to-3D methods still suffer low correspondence with the 2D input image, yielding unsatisfactory 3D output and modern text-to-3D methods also suffer low diversity in generated 3D objects from the same input text and high computation complexity.
To mitigate these limitations in 3D object generation, we propose DITTO-NeRF, a diffusion-based iterative text to omni-directional 3d model, which utilizes the high diversity and high fidelity images generated by the latent diffusion model in response to a given text prompt. Specifically, our DITTO-NeRF incorporates a monocular depth estimation model~\cite{ranftl2020towards} to predict the depth corresponding to the image and subsequently builds a high-quality partial 3D object for limited angles. NeRF is then trained using inpainting-SDS (Score Distillation Sampling)~\cite{rombach2021highresolution} loss for diffusion model to create an image corresponding to text prompt to fill the remaining part of the 3D representation. For better reconstruction of the 3D object in the early stage of training, we propose progressive global view sampling. Lastly, with the refinement stage, we could minimize discrepancies among generated parts. By utilizing these novel techniques, our method has the ability to construct 3D objects from text-generated images or given images.

We demonstrated the effectiveness of our method in both image-to-3D and text-to-3D tasks. 
In the image-to-3D task, our method outperformed those of the current state-of-the-art (SOTA) baselines in terms of both multi-view consistency and source image correspondence visually and in user studies.
As such, DITTO-NeRF's effectiveness is extended beyond that of previous models.
In the text-to-3D task, our method excelled those of the current SOTA baselines in terms of both output fidelity and diversity. Importantly, these improvements were achieved while requiring reasonable training time and computation resources. 
Here is the summary of our contributions:
 \begin{itemize}
\item Proposing a novel pipeline for generating a 3D object model from a single image or text prompt, called DITTO-NeRF, that iteratively propagates high-quality partial 3D model on in-boundary (IB) angles to the remaining 3D model on outer-boundary (OB) angles.
\item Proposing progressive global view sampling (PGVS) from IB to OB, reliability-guidance masking for IB, and multi-scale consistency refinement for all IB and OB. 
\item Outperforming prior arts on text/image-to-3D~\cite{xu2022sinnerf,jain2021putting, Xu_2022_neuralLift}, achieving remarkable results in terms of diversity/quality and speed/fidelity, respectively.
\end{itemize}

\begin{figure*}
\begin{center}
\includegraphics[width=2\columnwidth]{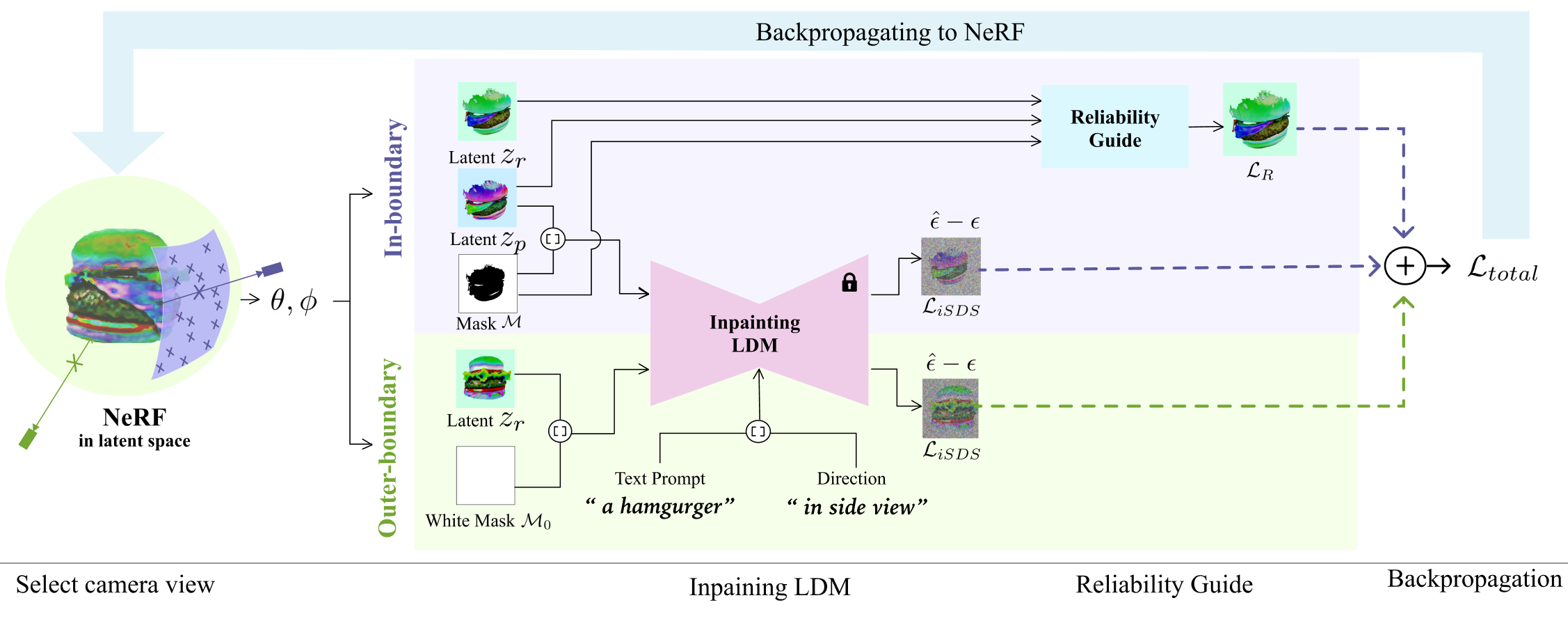}
\end{center}
   \caption{Our DITTO-NeRF pipeline for training from a 2D single image (given/generated). (Step 1) 2D latent $z_r$'s are sampled from latent NeRF at angles $\theta$ and $\phi$ using PGVS. (Step 2-1) Inpainting LDM with the given text and additional direction text yields the residual $\hat{\epsilon} - \epsilon$ from latent $z_r$ on OB or latent $z_p$ on IB. (Step 2-2) Latent $z_r$ on IB will be suppressed for outer-mask initially (so that latent $z_p$ will be relatively well-preserved for in-mask initially) and then progressively preserved for all area later with more reliable estimates.}
\label{fig:short}
\label{fig:pipeline}
\end{figure*}
 
\section{Related Works}

\subsection{Text-to-image guidance}
 In recent years, generative models have made significant strides in the field of image generation, an area that was once thought to be the exclusive domain of humans. Numerous studies have explored the use of generative models, including Variational AutoEncoder (VAE)~\cite{diederik2014auto, rezende2014stochastic, razavi2019generating}, flow-based models~\cite{dinh2014nice, dinh2016density}, and GAN~\cite{brock2018large, goodfellow2020generative, karras2019style}. However, the SOTA models are based on the denoising diffusion probabilistic model~\cite{dhariwal2021diffusion, song2020denoising, ho2022cascaded, ho2022classifier, ho2020denoising} (hereinafter referred to as the diffusion model). The diffusion model is comprised with forward diffusion steps that adds pre-defined noise based on time steps and reverse generative steps that denoise noisy images. These diffusion models have achieved exceptional image quality, but their applicability has been limited due to their high computational resource requirements.
  The latent diffusion model (LDM)~\cite{rombach2022high} has emerged as a promising approach for addressing the computational resource requirements associated with the forward/reverse process in image generation. This model leverages the encoding process of the VAE to process the forward and reverse processes in the latent space, rather than the RGB space, thereby significantly reducing the amount of computation required. By passing the final denoised latent vector through the VAE decoder, high-quality images can be obtained with reduced computational effort. The development of LDM has led to the emergence of several novel applications in image generation, such as inpainting models that incorporate additional inputs, such as masks, to manipulate the generated images based on the surrounding image context. 
  
  In this study, we leveraged an inpainting model based on LDM to achieve the goal of creating a 3D object that matches the user-input or generated image.

\subsection{Image-to-3D}
Various methods exist for representing 3D objects, including point clouds~\cite{achlioptas2018learning, luo2021diffusion, mo2019structurenet, yang2019pointflow, zhou20213d}, meshes~\cite{gao2022get3d, zhang2020image}, and voxels~\cite{gadelha20173d, smith2017improved, wu2016learning}. However, these methods require significant storage capacity to represent high-quality 3D objects. Neural Radiance Fields (NeRF)~\cite{mildenhall2020nerf} offers a novel solution for it by learning a neural network from images of certain viewpoints.
NeRF has been criticized for its long training times, rendering times, and dependency on a large number of camera positions and images. Consequently, much research has focused on optimizing training and rendering times~\cite{xu2022point, fridovich2022plenoxels, yu2021plenoctrees, muller2022instant, hu2022efficientnerf, deng2022depth, wang2021ibrnet, reiser2021kilonerf, garbin2021fastnerf, chen2021mvsnerf}. Instant-NGP~\cite{muller2022instant} using a multi-resolution hash grid encoding shows promising results. Additionally, several studies have explored the potential for achieving high-quality 3D representations with a limited number of images~\cite{deng2022depth, chen2021mvsnerf, yu2021pixelnerf, Xu_2022_SinNeRF, jain2021putting, Xu_2022_neuralLift} by leveraging the prior knowledge of pre-trained models. 

By combining the LDM approach with Instant-NGP, we were able to efficiently create a high-quality, multi-view consistent 3D representation from a single image.

\subsection{Text-to-3D}
In recent years, the diffusion model has shown outstanding performance in text-to-image tasks, and several attempts have been made to extend it to text-to-3D object tasks. DreamFusion has achieved remarkable results by leveraging that the noise residual value obtained from one step of the denoising generative process in the Diffusion Denoising Probabilistic Model~\cite{dhariwal2021diffusion} alters image to match the prompt.
The model adds Gaussian noise to the image rendered by NeRF in a forward process and backpropagates the noise residual obtained through the reverse process as a gradient to align the rendered image with the input text prompt. By performing this process iteratively for random views, the 3D object corresponding to the input text prompt can be obtained. 

In this paper, we propose DITTO-NeRF by adopting the main idea of DreamFusion. However, since DreamFusion uses Imagen~\cite{saharia2022photorealistic}, a proprietary diffusion model, we utilize Latent NeRF~\cite{metzer2022latent} which was based on Stable diffusion~\cite{rombach2022high} as the fundamental model.

\section{Methods}
 This chapter delineates the pipeline of our DITTO-NeRF, which comprises four distinct components: 1) The process of generating partial 3D objects from an image using LDM through an input text or a given image by a user is executed and the resulting objects are referred to as in-boundary objects ({IB 3D objects}). 2) NeRF employs progressive global view sampling (PGVS) and reliability-guided ($\mathcal{L}_{R}$) to effectively learn the frontal view, using the IB 3D object created through the aforementioned process. 3) $\mathcal{L}_{R}$ and inpainting SDS loss ($\mathcal{L}_{iSDS}$) are utilized to enhance the fidelity of the remaining frontal view. 4) A refinement stage is introduced to enhance the overall quality of the representation. The complete pipeline of DITTO-NeRF is presented in Fig.~\ref{fig:pipeline}.

\begin{figure}[!b]
\begin{center}\hspace*{-0.5cm}
\includegraphics[width=1.08\columnwidth]{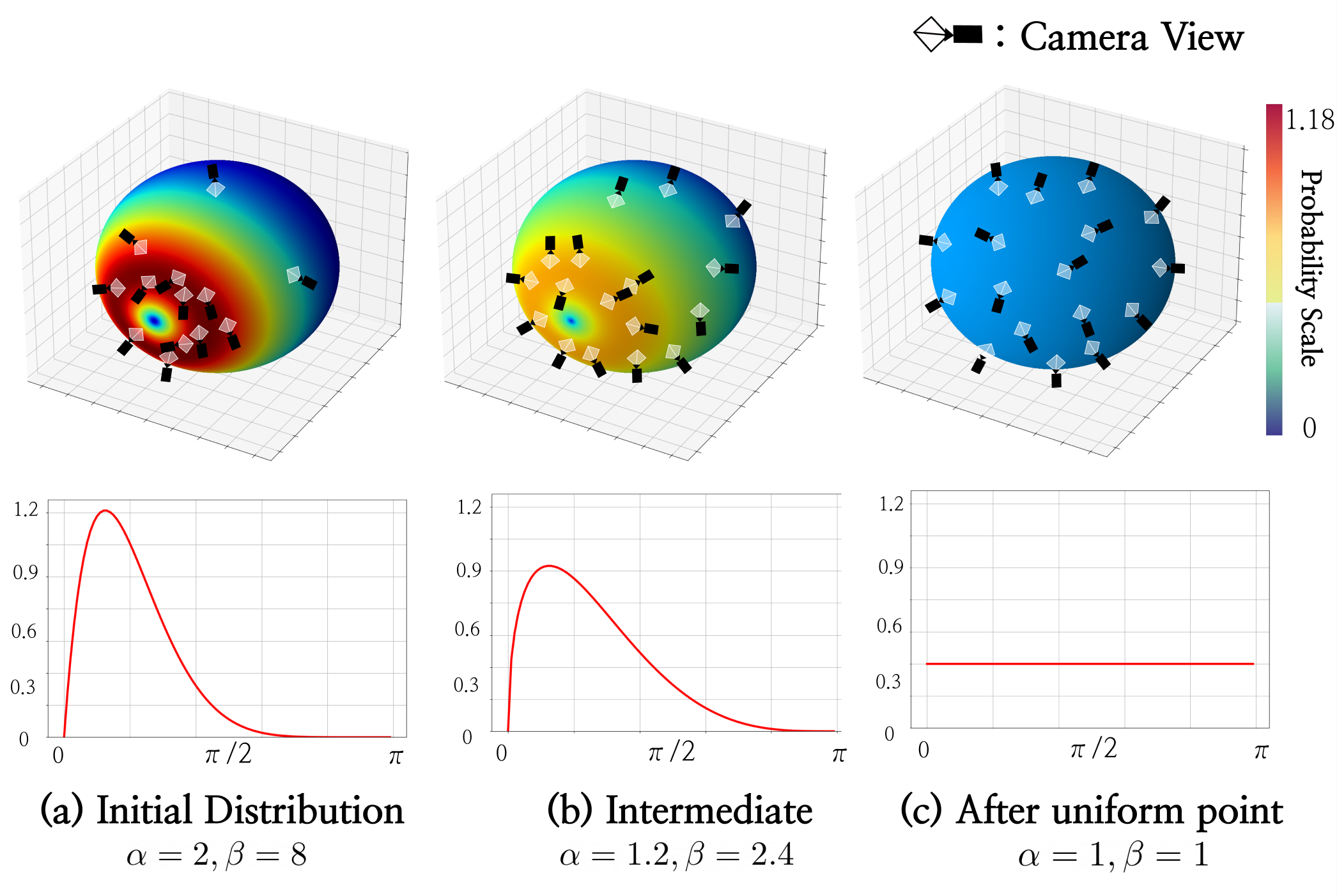}
\end{center}
   \caption{Examples of our PGVS over iterations based on varying Beta distribution. Initial camera view samples are heavily concentrated on IB (a) and then they are gradually spread to OB over iterations (b). Finally, camera view samples are uniformly distributed after uniform point.}
\label{fig:long}
\label{fig:sampling}
\end{figure}

\subsection{IB partial 3D object and pre-rendering}
 The diffusion model described previously can generate diverse images that correspond to a given text prompt. To further enhance the capabilities of the model for learning 3D representations that match the quality and diversity of the image generation model, we introduced a new process of creating a 3D object with a single image, and the object is named with {IB 3D object}. This process involves the creation of a partial 3D object that is utilized to aid NeRF in learning a 3D representation that aligns with the generated or user-given image.

\paragraph{IB 3D object.} A relative depth map was first extracted from the image using a monocular depth estimation model called MiDaS~\cite{ranftl2020towards}. Using this depth map, a 3D object can be composed with the form of a point cloud. It has been observed that in the point cloud constructed in this step, extraneous points such as backgrounds or floors may be inadvertently included in the point cloud. This can result in unwanted points being incorporated into the generated mesh during the conversion of the point cloud to mesh. To circumvent this, the method identifies outlier vertices based on their proximity to other vertices and subsequently removes them using a predetermined standard deviation. The values of these thresholds were determined heuristically via experimentation. Following this, Poisson surface reconstruction~\cite{kazhdan2013screened} is employed to generate the mesh from the completed point cloud.
After the Poisson surface reconstruction process, the 3D mesh of the desired object is acquired by removing the parts with less density than a certain quantile.

\paragraph{Setting IB and pre-rendering.} Rendering a 3D mesh in real-time for various viewpoints and using it for learning is computationally demanding. To mitigate this, the proposed method pre-renders the IB 3D object by sampling N views with uniform distribution within a limited angle range, which we call \textit{in-boundary} (IB). Specifically, RGB images, depth, and latent vectors are rendered using the VAE encoder of the LDM.

\subsection{Training NeRF using IB 3D object}

\paragraph{Progressive global view sampling (PGVS).} To enable the acquisition of the IB 3D object that was previously established and ensure coherence between the partial representation and subsequent portions generated through the diffusion model, it is imperative to initially sample numerous views of the in-boundary part. As the training progresses, views of the \textit{outer-boundary} (OB) part need to be sampled to facilitate the generation of the overall 3D object. To accomplish this, we employ the beta distribution as the probability density function to sample the viewing position. 
 \begin{equation}
 f(x; t) =  \twopartdef { \frac{x^{\alpha(t) - 1} (1 - x)^{\beta(t) - 1}}{B(\alpha(t), \beta(t))}  } {t < t_u} {U(0, 1)} {t \geq t_u}
 \end{equation}
 with $\alpha(t)=\alpha_0+1 - \frac{\alpha_0}{t_u}t $ and $ \beta(t)=\beta_0+1 - \frac{\beta_0}{t_u}t$.
 
  The point which is named the \textit{uniform point} $t_u$, is where $\alpha$ and $\beta$ decay to 1, after which uniform sampling is performed for all directions in the subsequent training steps which is illustrated in Fig.~\ref{fig:sampling}. Afterward, the process of obtaining the angle we want from the corresponding pdf is expressed by the following formula:
  \begin{equation}
 \theta, \phi = ( \bigintssss f(x;t) dx)^{-1}(U_{\theta, \phi}(0, 1)) \cdot range(\theta, \phi)
 \end{equation}
  If the sampled camera position lies within the in-boundary region, the nearest pre-rendered position is selected instead of rendering a random view each time to avoid the unnecessary computational burden. Since the camera view is sampled progressively, we call this sampling {Progressive global view sampling (PGVS).}

\subsection{Matching IB and OB 3D object}

\paragraph{Inpainting SDS loss.} Simultaneously applying the previously rendered prior images and the $\lambda_{iSDS}$ from the diffusion model can result in a conflict that depreciates the consistency of the IB 3D object and $\mathcal{L}_{iSDS}$ generated parts as the diffusion model may prioritize creating an object that corresponds to the built-in prior model. This can impede the goal of generating a 3D object that closely resembles the desired image. To address this, we employ a fine-tuned diffusion model $\epsilon_{\phi}$~\cite{rombach2022high} for the inpainting task, rather than using a general diffusion model. 
\begin{equation}
    \mathcal{L}_{iSDS}(\phi, g(\theta)) =  \mathbb{E}_{t, \epsilon}[||\epsilon_{\phi}(x_t ; y, t, \mathcal{M}) - \epsilon_t||_2^2]
\end{equation}
where $y$ is text embedding, $\mathcal{M}$ is the binary mask, $\epsilon_t$ is actual noise for time step $t$, and $x_t$ is noise image for each time step $t$.
When the sampled viewpoint is in-boundary, the rendered latent vectors and pre-rendered latent images are optimized with $\mathcal{L}_{R}$. In addition, sparsity loss ($\mathcal{L}_{sp}$) was introduced, which was to obtain a cleaner representation by suppressing the lump caused by learning NeRF. For further details for $\mathcal{L}_{sp}$, see the supplementary material. The total loss is formulated as shown in the accompanying equation.
\begin{equation}
\mathcal{L}_{total} = \lambda_{iSDS}\mathcal{L}_{iSDS} + \delta_{R}\left(\theta, \phi\right)\mathcal{L}_{R} + \lambda_{sp}\mathcal{L}_{sp}
\end{equation}
$\delta_{R}(\theta, \phi)$ here has a value of 1 within the IB area and a value of 0 otherwise. 

\paragraph{Reliability-guided loss.} When the camera position is sampled in the IB area, we optimize the loss between the rendered latent image in NeRF and the pre-rendered latent image. However, due to the difference in consistency and size between the IB 3D object and the generated object, we introduce a novel loss, which takes into account the differences between the part of the image corresponding to the desired object (foreground) and the remaining region (background), called the reliability-guided loss.
\begin{equation}
\mathcal{L}_{R} = [\zeta \cdot \mathcal{M} + \eta(t) \cdot (1 - \mathcal{M})] \odot || z_{r} - z_{p} ||_{1}
\end{equation}
with $\eta(t) = e^{-t/\lambda_\eta}$. $\lambda_{\eta}$ and $\zeta$ are constants that found with experiments. $z_{r}$ is rendered latent vector from NeRF and $z_{p}$ is pre-rendered latent vector.

The background of pre-rendered images is initialized as white. However, learning the latent ground truth as it leads to suboptimal inpainting results in the IB region, as the $\mathcal{L}_{iSDS}$ is smaller than the $\mathcal{L}_{R}$. On the other hand, if the diffusion model generates the background and the camera position is sampled from the back of the object, the model may not recognize the IB 3D object, resulting in inconsistent objects with varying sizes. Therefore, to suppress the unreliable background area at the beginning of training, we also train the white part of the background at the beginning. After that, learning proceeds, and since the IB 3D object is sufficiently represented in NeRF, the part generated by the $\mathcal{L}_{iSDS}$ becomes reliable, and the weight of the part against the background is exponentially reduced to generate the rest part.

\begin{figure}[t]
\begin{center}
\includegraphics[width=1.04\columnwidth]{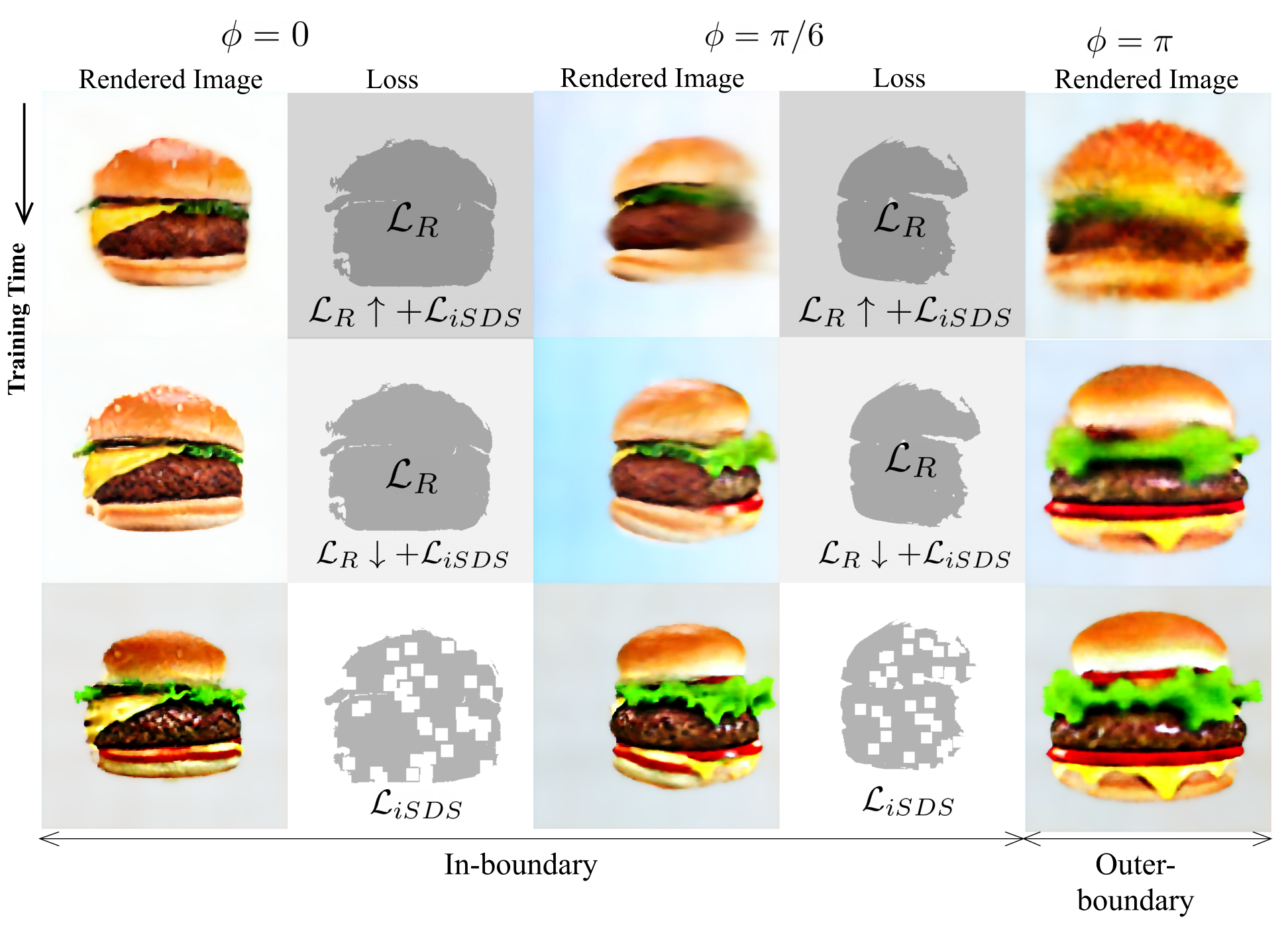}
\end{center}
   \caption{This figure illustrates how does the $\mathcal{L}_{R}$ and refinement works. At the beginning of the training process, $\mathcal{L}_{R}$ suppresses the unreliable part so that IB 3D object could be represented. As the training proceeds reliability of $\mathcal{L}_{iSDS}$ increases creating outer-boundary parts. In refinement procedure, random patches are applied to mask for enhancing overall quality.}
\label{fig:long}
\label{fig:mask}
\end{figure}

Specifically, during the calculation of the $\mathcal{L}_{R}$, the loss is divided into {foreground} and {background} parts, with the foreground part given a higher weight and the background part given a relatively lower weight to initially create a white background. Subsequently, during training, the {background} weight is gradually decayed, allowing the $\mathcal{L}_{iSDS}$ for inpainting to have a more significant impact if the in-boundary region is learned after a certain amount of training. Through this process, we were able to train a NeRF model that created a 3D object that resembles an image prior with a constant size. This whole process is illustrated in Fig.~\ref{fig:mask} along with rendered training images for each step.

\begin{figure*}
\begin{center}\vspace{-0.7cm}
\includegraphics[width=1.9\columnwidth]{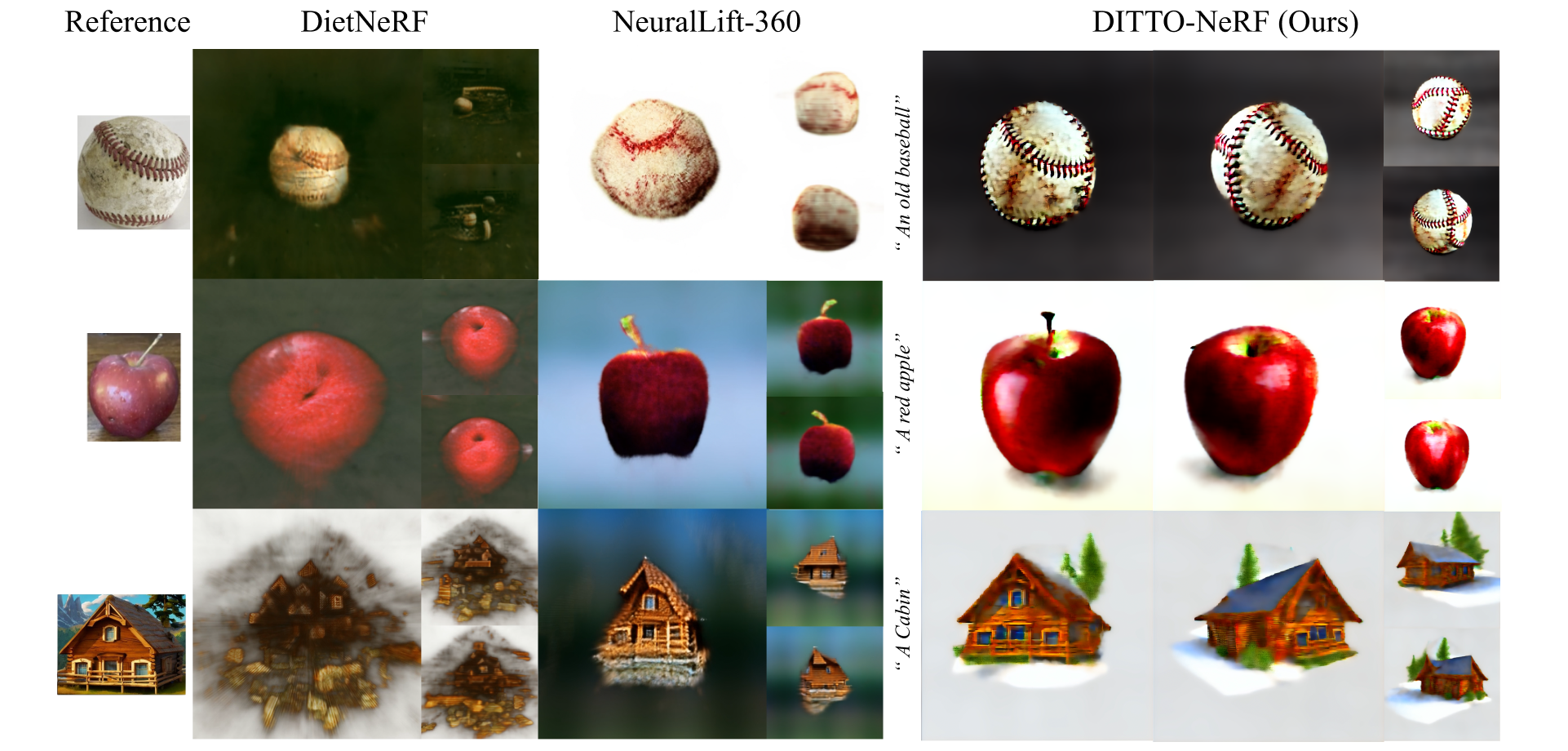}
\end{center}
\vspace{-0.6cm}
   \caption{Qualitative comparison with other image-to-3D models~\cite{jain2021putting, Xu_2022_neuralLift}. The outputs of the Neurallift-360~\cite{Xu_2022_neuralLift} were obtained and cropped from the Neurallift-360 website for fair comparison. Our model used the reference images and the corresponding texts on the left.}
\label{fig:short}
\label{fig:comparison}
\end{figure*}

\begin{figure*}
\begin{center}\vspace{-1em}
\includegraphics[width=1.92\columnwidth]{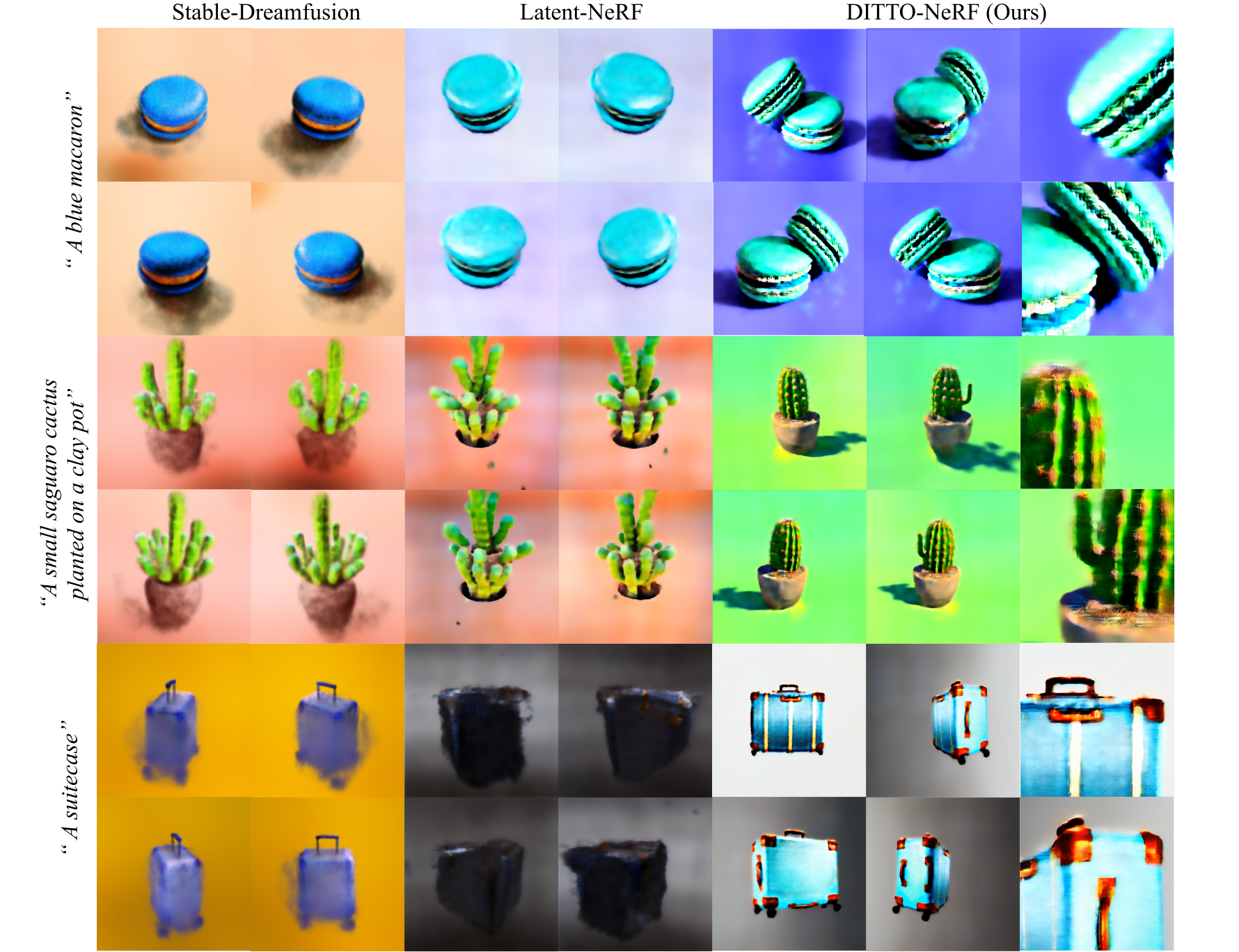}
\end{center}
\vspace{-0.4cm}
   \caption{Qualitative comparison with other text-to-3D models~\cite{stable-dreamfusion, metzer2022latent}. The last column is our model's zoomed-in results to show our excellent details. The number of iterations for the text-to-3D model to generate outputs was set to the default value in the baseline.}
\label{fig:short}
\label{fig:comparison}
\end{figure*}

\begin{figure*}
\begin{center}
\includegraphics[width=2\columnwidth]{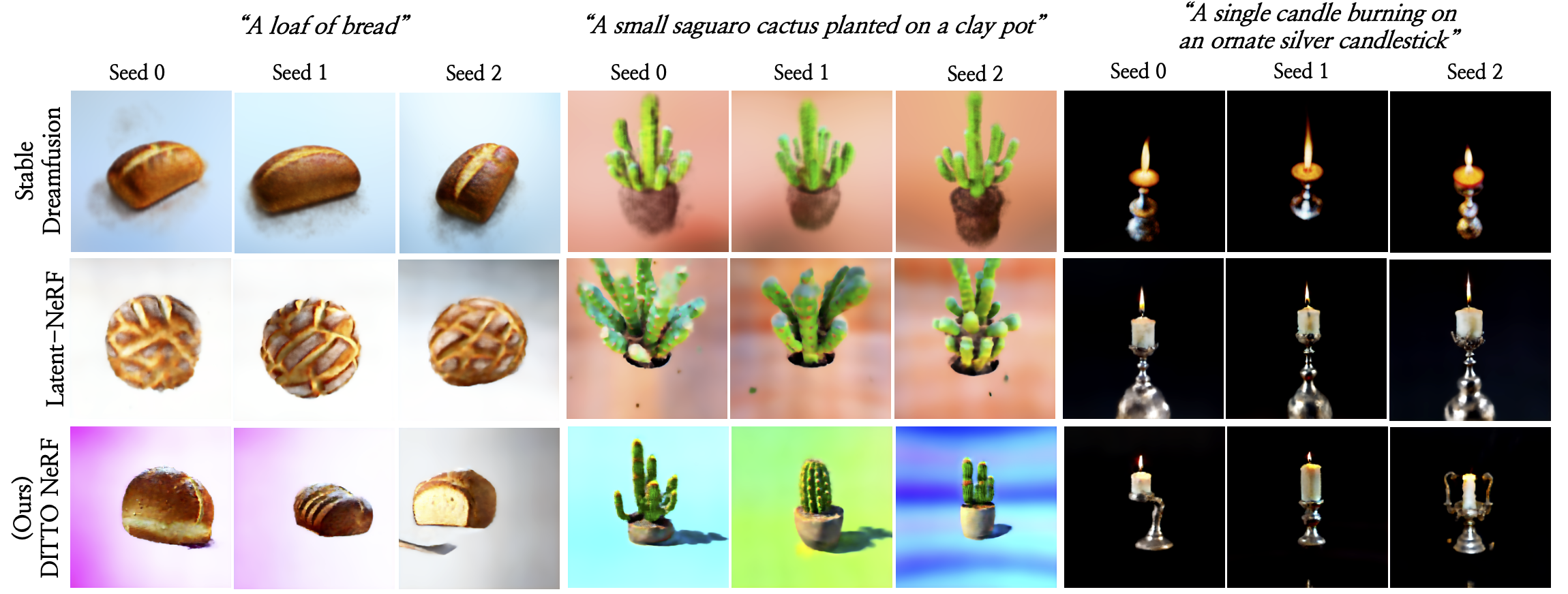}
\vspace{-0.4cm}
\end{center}
   \caption{Qualitative comparison with other text-to-3D models~\cite{stable-dreamfusion, metzer2022latent} in terms of diversity. All seeds were taken randomly.}
   \vspace{-0.2cm}
\label{fig:short}
\label{fig:diversity}
\end{figure*}

\subsection{Refining details}

\paragraph{Random patch refinement.} Although the part generated by the $\mathcal{L}_{iSDS}$ and IB 3D object part continue semantically seamless, there is a noticeable discontinuity in the overall texture and color. This is attributed to the prior inside the diffusion model. To address this challenge and enable the created 3D object to exhibit continuity of texture and color while retaining the image prior of the front part as much as possible, a \textit{refinement step} was introduced. This refinement step constitutes 10\% of the entire training process, and if in-boundary is sampled during this step, the $\mathcal{L}_{R}$ is excluded, and a random patch is given to the foreground to refine the corresponding part with $\mathcal{L}_{iSDS}$. This process results in the color and texture of the generated 3D object becoming more similar to the part generated by the $\mathcal{L}_{iSDS}$, while maintaining the overall shape and content of the image prior. Through the refinement step, the desired objective of achieving continuity of texture and color while maintaining the image prior is attained. The corresponding method is shown in Fig.~\ref{fig:mask}

\paragraph{Dimension refinement.} In addition, it was observed that the pipeline resulted in the appearance of jagged edges in the generated object, known as the \textit{jaggies} phenomenon. This issue arises due to the training process, which involves feeding a relatively high-resolution image into a relatively low-resolution latent space. To address this, we introduced a refinement step that linearly doubles the NeRF rendering resolution, resolving the issue of jagged edges. Through this process, we were able to generate an object with clear edges while maintaining the overall shape and content of the IB 3D object. The color and texture of the generated object were also found to be similar to those produced by the $\mathcal{L}_{iSDS}$. Check the supplementary for the details.

\begin{table}[] 
\begin{center}
\begin{tabular}{c|cc} 
\hline
Method         & Corr. $\uparrow$ & Fidelity $\uparrow$ \\ \hline\hline
DietNeRF         & 2.29                   & 2.194        \\
SinNeRF           & 2.084                  & 2.031        \\
NeuralLift-360    & 3.221                  & 3.24        \\
\begin{tabular}[c]{@{}c@{}}\textbf{DITTO-NeRF} \\ (\textbf{Ours})\end{tabular} & \textbf{3.928}                  & \textbf{4.019}        \\ \hline
\end{tabular}
\end{center}
\vspace{-0.2cm}
\caption{Mean opinion score for evaluating image-to-3D models.}
\vspace{-0.4cm}
\label{table:Image-to-3D}
\end{table}

\section{Experiments}
In this section, we aim to evaluate the effectiveness of our proposed method and compare it with other existing models in the domains of image-to-3D and text-to-3D. For image-to-3D, we perform a comparative analysis between our method and existing models with respect to source-generated 3D object correspondency and fidelity. Similarly, for text-to-3D, we compare the performance of our method with other open-sourced models in terms of diversity, computational efficiency, and fidelity. Additionally, we investigate the impact of various factors that we introduced on method section. The user study's respondents were asked to rate each item on a 5-point scale. 3150 questionnaires were collected from a total of 210 people. All subsequent experiments were done on a single NVIDIA A100 GPU.

\subsection{Generating 3D objects from a single image}
 In the present section, a comparative analysis is carried out between our proposed method and existing single-image NeRF models. To evaluate the effectiveness of our approach, a questionnaire survey is conducted to obtain N responses on source-generated object correspondence and fidelity. The survey responses provide insight into the performance of our method relative to the SOTA models in terms of its ability to accurately reconstruct objects and maintain correspondence with the source. 

 \paragraph{User study.} A survey was carried out to evaluate the effectiveness of our proposed method in comparison to existing image-to-3D generative models, namely SinNeRF~\cite{xu2022sinnerf}, DietNeRF~\cite{jain2021putting} and Neurallift-360~\cite{Xu_2022_neuralLift}. The survey was designed to gather user feedback on the fidelity of the generated object and source image-object correspondency. The results in Table~\ref{table:Image-to-3D} indicate that our method outperforms the existing models in both categories, establishing a new SOTA in image-to-3D generative models.


\begin{table}[]
\begin{center}
\begin{adjustbox}{width=\linewidth}
\begin{tabular}{c|cccc}
\hline
Method
& Diversity $\uparrow$ & Fidelity$\uparrow$ & Corr.$\uparrow$ & Time (m) $\downarrow$ \\ \hline \hline
\begin{tabular}[c]{@{}c@{}}Stable\\ DreamFusion\end{tabular} & 3.704     & 2.6      & 2.959     & 60    \\
Latent-NeRF                                                   & 2.995     & 3.094    & 3.362     & \textbf{10}    \\
\begin{tabular}[c]{@{}c@{}}\textbf{DITTO-NeRF}\\ (\textbf{Ours})\end{tabular}  & \textbf{3.966 }    & \textbf{4.158}    & \textbf{4.242}     & 25    \\ \hline
\end{tabular}
\end{adjustbox}
\end{center}
\vspace{-0.2cm}
\caption{Mean opinion score and training time for evaluating text-to-3D models.}
\vspace{-0.4cm}
\label{table:Text-to-3D}
\end{table}

\begin{figure*}
\begin{center}
\includegraphics[width=2\columnwidth]{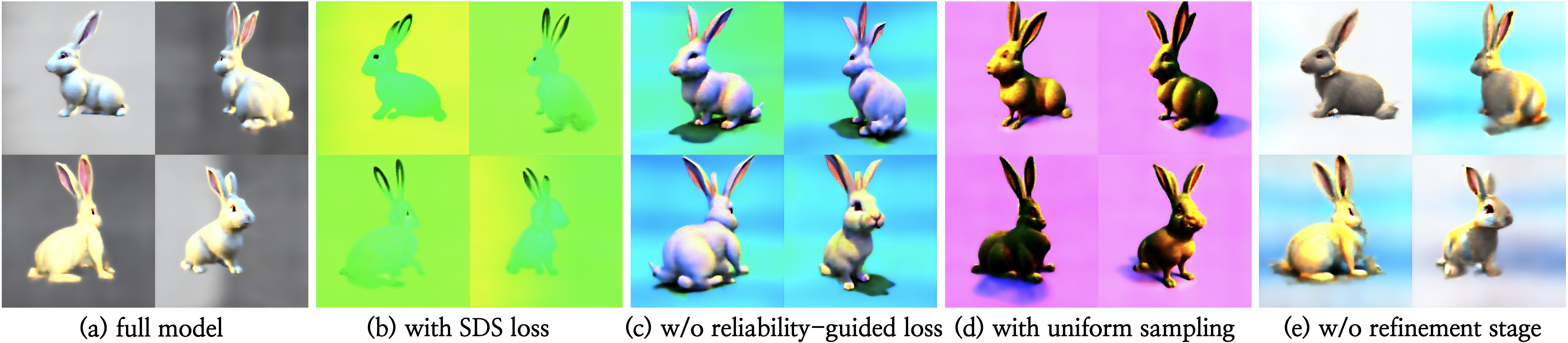}
\end{center}
   \caption{Ablation studies by generating 3D object with a text prompt \textit{``a rabbit, animated film character, 3D rendered"} using full model (a), without i-SDS loss (b), without reliability-guided loss (c), with uniform sampling (not PGVS) (d), and without refinement stage (e).}
\label{fig:short}
\label{fig:ablation}
\end{figure*}

 \subsection{Generating 3D objects from a text prompt}
 This section presents a comparative analysis of our proposed method with existing Stable diffusion-based text-to-3D generative models~\cite{stable-dreamfusion, metzer2022latent}. The evaluation is conducted through a questionnaire survey that captures respondents' feedback on the diversity and fidelity of the generated 3D objects. The survey responses are scored on a 5-point scale for each item. Furthermore, to investigate the computational efficiency of the models, we report the time spent on the baseline's default setting.

 \paragraph{Diversity with a single prompt.} Table~\ref{table:Text-to-3D} presents the results of our comparative analysis, indicating that our proposed method exhibits higher diversity in generated 3D objects as compared to the Stable diffusion-based text-to-3D generative models. An example of this diversity is illustrated in Fig.~\ref{fig:diversity}, where multiple outputs are generated for some given prompts. 

 \paragraph{Fidelity and text correspondency.} As observed in previous studies, the Stable diffusion-based text-to-3D generative models produce unrealistic, blurry, or unclear 3D objects. In contrast, our proposed method generates relatively realistic 3D objects, as evidenced by the results presented in Table~\ref{table:Text-to-3D}, supported by users. Similarly, the correspondence between the input text and the generated object is observed to be more accurate and consistent in our method, owing to its superior performance in accurately capturing the semantic meaning of the input text. 

\paragraph{Computational efficiency.} Stable-DreamFusion which is based on DreamFusion, suffered from the issue of requiring a long training time to achieve results with satisfactory quality. Latent-NeRF, on the other hand, addressed this issue by training the model on the latent space; however, it exhibited limitations in terms of diversity and fidelity. In contrast, our proposed method is capable of learning a 3D representation of satisfactory quality within a reasonable training time, as illustrated in Table~\ref{table:Text-to-3D}. While our method may have some limitations in fidelity when compared to models such as Magic3D or DreamFusion, it should be noted that these models utilize large and undisclosed models, limiting their applicability. In contrast, our proposed method offers an advantage in terms of computation.

 \subsection{Ablation study}
In this section, we conduct an ablation study on the components of our method. We used inpainting SDS loss instead of conventional SDS loss to make the IB 3D object and diffusion-generated part continued smoothly. Otherwise, as shown in Fig.~\ref{fig:ablation}, it was obvious that the shape and color were lost. In the case where $\mathcal{L}_{R}$ does not exist, it was confirmed that the generated object was larger than IB 3D object or additional elements were added to it. There was a problem that the 3D prior could not be reconstructed when sampling viewpoints uniformly without using decaying beta distribution. Finally, if refinement was not performed, the semantic part continued, but discontinuity occurred in color and texture. Further detailed explanation is on the supplementary.

\section{Limitation}
\begin{figure}[t] 
\begin{center}
\includegraphics[width=1\columnwidth]{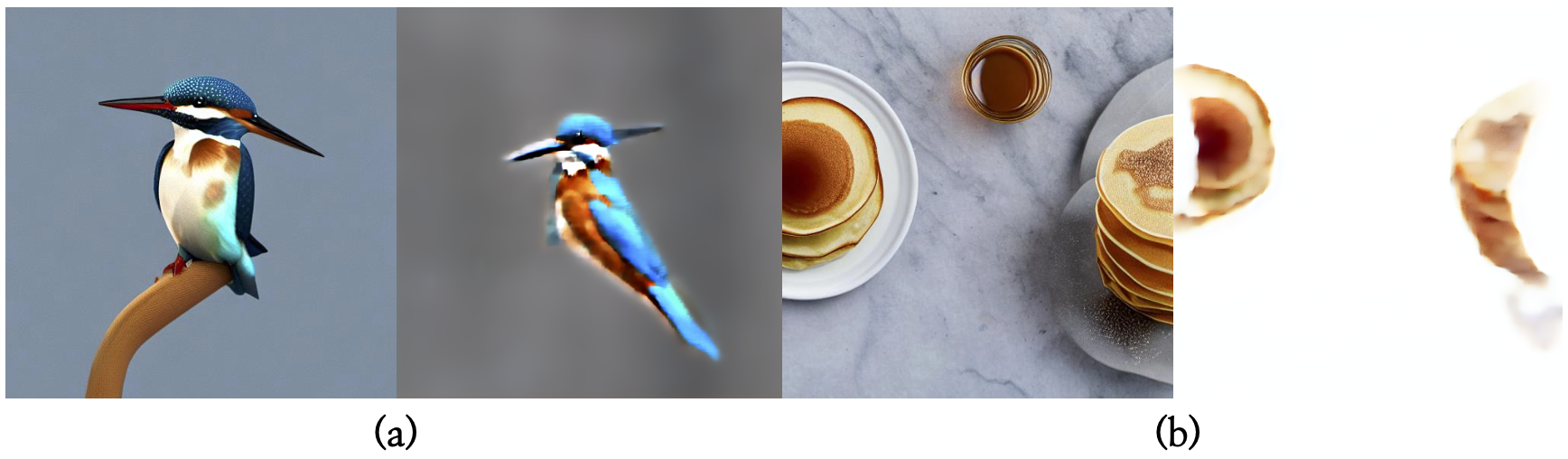}
\end{center}
   \caption{As the diffusion model failed to generate proper images, the generated 3D objects' quality are also limited.}
\label{fig:long}
\label{fig:limitation}
\end{figure}

Depth prediction is based on the shadows present in the image and the prior of the model. Therefore, if an image with minimal shadows or an image that is generated from outside of the distribution of the diffusion model is input, accurate depth estimation may not be possible. This can result in an inadequate IB 3D object being generated, leading to lower-quality 3D object outputs compared to those obtained in general. Additionally, if the quality of the image generated by the diffusion model is poor, learning can become more challenging as illustrated in Fig.~\ref{fig:limitation}. Also, in cases where a low-probability image is selected from the diffusion model, it may not follow the image prior, leading to a convergence of the 3D object within the diffusion model. This can also cause a discontinuity issue between the image prior and diffusion-generated parts.


\section{Conclusion}
We introduce DITTO-NeRF, a novel approach for obtaining a 3D representation from a single image or text input. DITTO-NeRF leverages an IB 3D object built from images obtained from user input or text, which is then used to train a NeRF model that generates 3D objects using a diffusion model. Based on user studies and qualitative comparisons, we draw the following conclusions: our proposed model achieves higher fidelity and quality in terms of image-to-3D correspondence than existing works. Moreover, in the context of text-to-3D generation, our method offers both higher fidelity and more computation-efficient diversity than existing Stable diffusion-based studies.

\section*{Acknowledgements}
This work was supported by the National Research Foundation of Korea(NRF) grants funded by the Korea government(MSIT) (NRF-2022R1A4A1030579) and Basic Science Research Program through the NRF funded by the Ministry of Education(NRF-2017R1D1A1B05035810).



\iccvfinalcopy 

\def\iccvPaperID{XXXX} 
\def\httilde{\mbox{\tt\raisebox{-.5ex}{\symbol{126}}}}

\clearpage
\appendix

\twocolumn[{%
\renewcommand\twocolumn[1][]{#1}%

\begin{center}
\bigskip 
\bigskip 
\textbf{\Large DITTO-NeRF: Diffusion-based Iterative Text To Omni-directional 3D Model \\ (Supplementary Material) \\}

\bigskip 
\bigskip 
\maketitle
 
\end{center}%
}]

\setcounter{equation}{0}
\setcounter{figure}{0}
\setcounter{table}{0}
\setcounter{page}{1}
\makeatletter
\renewcommand{\theequation}{S\arabic{equation}}
\renewcommand{\thefigure}{S\arabic{figure}}
\renewcommand{\thetable}{S\arabic{table}}

\section{Additional Results}
\subsection{Videos}

We present a video showcasing a comparison of the results obtained using our proposed DITTO-NeRF model, which is a novel method for learning 3D representations from either image or text inputs, against other baselines in the same task. For video, please check the folllowing page: \href{https://janeyeon.github.io/ditto-nerf}{{\texttt{janeyeon.github.io/ditto-nerf}}}.

\section{Details on Methods}
\label{app1_methods}
\subsection{Algorithms}
\paragraph{Overall pipeline.}

\begin{algorithm}[h!]
    \caption{Overall pipeline\label{algo_overall}}
    \DontPrintSemicolon
    \SetAlgoNoLine
    \SetAlgoVlined
    \SetKw{Continue}{continue}
    \SetKw{Break}{break}
    \SetKwFunction{train}{Training}
    \SetKwFunction{refine}{Refinement}
    \SetKwFunction{find}{Find-closest}
    \SetKwFunction{generate}{Generate-IB-3D}    \SetKwFunction{render}{Render-ray}
    \SetKwFunction{pgvs}{PGVS}
    \SetKwFunction{sparsity}{Sparsity}
    \SetKwFunction{reliable}{Reliability-Guide}
    \SetKwFunction{dim}{Dimension-Refine}
    \SetKwFunction{add}{Add-Patch}
    \SetKwProg{Fn}{Require}{:}{}
    \KwIn{$y$, $I$}
    \KwOut{$\mathbf{V}_{out}$}
     \tcp{Training \& Refinement}
   \uIf{$\exists!\ I$}{
    $y_{image} = CLIP((y, y_{bg}))$ \;
    $z_{image} \gets \textnormal{inpainting-LDM}(y_{image})$ \;
    $I \gets \mathcal{D}(z_{image})$
   } 
   $\mathbb{Z}_p, \mathbb{M} \gets \generate(I)$ \;

    \For{i $= 1, 2, ..., t_{total}$}{
    $\theta, \phi \gets \pgvs(i)$ \;
    \uIf{$IB$}{
        $z_p,\mathcal{M} \gets \find(\mathbb{Z}_p, \mathbb{M}, \theta, \phi)$\;
    }
    $\mathbf{o}, \mathbf{d} \gets \render(\theta, \phi)$ \;
    $z_r,\  \mathcal{P}_{ws} \gets \mathcal{N}(\mathbf{o}, \mathbf{d})$ \;

    \uIf{$i \le i_{refine}$}{
        $\train(\mathcal{N}, z_p, \mathcal{M}, z_r, \mathcal{P}_{ws}, y)$
    } \Else{
        $\refine(\mathcal{N}, z_p, \mathcal{M}, z_r, \mathcal{P}_{ws}, y, i)$
    }
    }
     \tcp{Rendering output} 
    $\mathbf{V}_{out}  \gets \{\}$ \;
    \For{$ \theta, \phi \textnormal{ in rendering angles}$}{
    $\mathbf{o}, \mathbf{d} \gets \render(\theta, \phi)$ \;
    $z_r \gets \mathcal{N}(\mathbf{o}, \mathbf{d})$ \;
    $I_{out} \gets \mathcal{D}(z_r)$ \;
    $ \textnormal{Append }I_{out} \textnormal{ to }\mathbf{V}_{out}$
    }
\end{algorithm}

The overall algorithm of DITTO-NeRF is described in Algorithm~\ref{algo_overall}. First, it requires text input $y$ and additional image input $I$ from the user. In the case of image-to-3D synthesis, the algorithm receives a reference image $I$ and corresponding text value $y$ as input. Whereas in the case of text-to-3D synthesis, only text input is required. If no reference image is available, two text values $y, y_{bg}$ are concatenated to create a text embedding $y_{image}$ that is used to generate a latent vector $z_{image}$ in inpainting-LDM. $z_{image}$ is then passed through a decoder $\mathcal{D}$ to create a reference image $I$. $I$ is fed into the $\generate$ function to obtain latent vector $z_p$ and mask $\mathcal{M}$  for randomly selected $N$ angles through uniform distribution within the in-boundary (IB) area.

For $N_{total}$ iterations, the algorithm performs the following procedures: the camera view position, $\theta$, and $\phi$ are determined using the $\pgvs$ function along the iteration. The IB flag is then used to determine whether the angle values are used directly or switched to pre-selected angles within the in-boundary area. Using the $\find$ function, $z_p$ and $\mathcal{M}$ values are obtained for each pre-selected angle. The $\render$ function is used to obtain the $\mathbf{o}$ and $\mathbf{d}$ values for each ray based on the $\theta$ and $\phi$. These values are used by the NeRF model ($\mathcal{N}$) to render the latent $z_r$ and weight values $\mathcal{P}_{ws}$, respectively, for each point of the ray. If the current iteration $i$ is less than $i_{refine}$, $\train$ is performed, and if it is larger, $\refine$ is executed.

As the $\mathcal{N}$ is trained, the latent vector $z_r$ is extracted from the $\mathcal{N}$ along the angles corresponding to `rendering angles'.
At each rendering angle, the rendered image $\mathcal{D}(z_r)$ is obtained from the latent vector $z_r$ using the decoder $\mathcal{D}$. By collecting rendered images $I_{out}$, we can obtain the final results,  $\mathbf{V}_{out}$, a video of a 3D rendered object.
 It should be noted that the output of the $\mathcal{N}$ is a 4-channel latent vector, unlike the RGB output in standard NeRF. So the $z_r$ should be passed through the $\mathcal{D}$ used by inpainting-LDM in the last rendering step to get an RGB image.

\begin{algorithm}[h!]
    \caption{Training\label{algo_train}}
    \DontPrintSemicolon
    \SetAlgoNoLine
    \SetAlgoVlined
    \SetKw{Continue}{continue}
    \SetKw{Break}{break}
    
    \SetKwProg{Fn}{Function}{:}{}
    \KwIn{$\mathcal{N}, z_p, \mathcal{M}, z_r, \mathcal{P}_{ws}, y$}
    \KwOut{$\mathcal{N}$}
    
    \Fn{\reliable{$z_p, z_r, \mathcal{M}$}}
    {
    \KwRet{$[\zeta \cdot \mathcal{M} + \eta(t) \cdot (1 - \mathcal{M})] \odot || z_{r} - z_{p} ||_{1}$}
    } 
    $y_{train} \gets CLIP((y, y_{dir}))$ \;
   \uIf{$IB$}{
    $\mathcal{L}_{iSDS} \gets \textnormal{inpainting-LDM}(z_p, y_{train}, \mathcal{M})$\;
    $\mathcal{L}_{R} \gets \reliable(z_p, z_r, \mathcal{M})$
   } \Else {
    $\mathcal{M}_0 \gets J_{H \cdot f_{scale}}$ \;
    $\mathcal{L}_{iSDS} \gets \textnormal{inpainting-LDM}(z_r, y_{train}, \mathcal{M}_0)$
   }

   $\mathcal{L}_{sp} \gets \sparsity (\mathcal{P}_{ws}) $\;
   
    $\mathcal{L}_{total} \gets \lambda_{iSDS}\mathcal{L}_{iSDS} + \delta_{R}\left(\theta, \phi\right)\mathcal{L}_{R} + \lambda_{sp}\mathcal{L}_{sp}$ \;
    $\textnormal{Update } \mathcal{N} \textnormal{ with } \mathcal{L}_{total}$
\end{algorithm}

\paragraph{Training procedure.}
As described above, when the current iteration $i$ is less than $i_{refine}$, $\train$ is performed. The entire algorithm of the $\train$ stage is described in Algorithm~\ref{algo_train}.  During this stage, the input text $y$ is concatenated with direction information corresponding to the angle and put into a CLIP to get $y_{train}$. The subsequent process differs depending on whether it is IB or not. In the case of IB, the $\mathcal{L}_{iSDS}$ is calculated by inputting the $z_p$ and $\mathcal{M}$ values into the inpainting-LDM. The difference between $z_r$ and $z_p$ is then corrected using the $\reliable$ function, which adjusts the background and foreground values using a $\mathcal{M}$ and expands the size of the reliable region as learning proceeds.
When it is not IB, network training is performed using the inpainting-LDM with $z_r$ and $\mathcal{M}_0$ values. After obtaining the $\mathcal{L}_{sp}$ to eliminate the lump, backpropagation is performed by adding all of the losses into $\mathcal{L}_{total}$.

 \begin{algorithm}[h!]
    \caption{Refinement\label{algo_refine}}
    \DontPrintSemicolon
    \SetAlgoNoLine
    \SetAlgoVlined
    \SetKw{Continue}{continue}
    \SetKw{Break}{break}
    \SetKwFunction{sparsity}{Sparsity}
   
    \SetKwProg{Fn}{Function}{:}{}
    \KwIn{$\mathcal{N}, z_p, \mathcal{M}, z_r, \mathcal{P}_{ws}, y, i$}
    \KwOut{$\mathcal{N}$}
    
    \Fn{\dim{$iter$}}
    {
    $rate \gets (iter - t_{total} \cdot f_{ref})/t_{total} \cdot (1 - f_{ref})$ \;
    $H, W \gets 64 \times (1 + rate), 64 \times (1 + rate)$ \;
    \KwRet{$H, W$}
    } 

    \Fn{\add{$\mathcal{M}$}}
    {
        \For{
            i = 0, 1, ..., $N_{patch}$
        }{
            $P_{x} = \textnormal{random}(S_{mask})$\;
            $P_{y} = \textnormal{random}(S_{mask})$\;
            $\mathcal{M}[P_{x}: P_{x} + S_{patch}, P_{y}: P_{y} + S_{patch}] \gets 1$
        }
    \KwRet{$\mathcal{M}$}
    } 
    $y_{refine} \gets CLIP((y, y_{dir}))$ \;
    $H, W \gets \dim(i)$ \;
   \uIf{$IB$}{
    $\mathcal{M} \gets \textnormal{resize}(H, W)$ \;
     $\mathcal{M} \gets \add(\mathcal{M})$ \;
    $z_p \gets \textnormal{resize}(H, W)$ \;
    $\mathcal{L}_{iSDS} \gets \textnormal{inpainting-LDM}(z_p, y_{refine}, \mathcal{M})$
   } \Else {
   $\mathcal{M}_0 \gets J_{H \cdot f_{scale}}$ \;
    $\mathcal{L}_{iSDS} \gets \textnormal{inpainting-LDM}(z_r, y_{refine}, \mathcal{M}_0)$
   }

   $\mathcal{L}_{sp} \gets \sparsity (\mathcal{P}_{ws}) $

   $\mathcal{L}_{total} \gets \lambda_{iSDS}\mathcal{L}_{iSDS} + \lambda_{sp}\mathcal{L}_{sp}$ \;
  $\textnormal{Update } \mathcal{N} \textnormal{ with } \mathcal{L}_{total}$
\end{algorithm}

\paragraph{Refinement procedure.}
When the current iteration $i$ is greater than $i_{refine}$, a refinement process is performed as outlined in Algorithm~\ref{algo_refine}. Unlike the previous training step, the $\reliable$ process is omitted, and the $\dim$ and $\add$ processes are added. The process of obtaining $y_{refine}$ is the same as the previous step, and the $\dim$ function calculates $H$ and $W$ based on the current iteration $i$. This function linearly increases the size of $H$ and $W$ from 64 to 128 as the refinement process progresses, for improvement of the resolution and quality in rendered views. To match the corresponding sizes of $H$ and $W$, the pre-calculated $\mathcal{M}$ and $z_p$ sizes are adjusted before entering. The $\add$ process enters only the $\mathcal{M}$ in the IB area and maintains consistency of the front and back sides by randomly arranging $S_{patch}$ size of $N_{patch}$ patches in a mask of $S_{mask}$ size. Subsequently, $\mathcal{L}_{iSDS}$ and $\mathcal{L}_{sp}$ are added to $\mathcal{L}_{total}$, and the $\mathcal{N}$ is trained in the same way as in the training process.

\subsection{IB 3D object pre-rendering}
To pre-render an IB 3D object, the angle of the in-boundary (IB) must be established first. In instances where this angle is too narrow, the matching between the IB and outer-boundary (OB) is insufficient. Whereas if the angle is too wide, additional views must be sampled, which can cause the bottleneck problem of $\find$ process in Algorithm~\ref{algo_overall}. Through several experiments, we discovered the setting $\phi \in [-30, 30]$ and $\theta \in [60, 120]$ provides optimal conditions. Moreover, if the number of pre-rendered images is too small, IB training cannot be performed sufficiently, while a large number of images increases the computational burden. Thus, based on our experimental findings, we determined that setting N=64 provides optimal conditions.

\subsection{Camera intrinsic parameters for point cloud}
In order to generate a point cloud from an RGB-D image, it is essential to know the camera intrinsic parameters. For real-world cameras, these values are readily available; however, as our images are synthesized using a diffusion model, the actual parameters remain unknown. The required intrinsic parameters include skew rate, $c_x$, $c_y$, $f_x$, and $f_y$. Given that contemporary cameras exhibit minimal skew, we approximated the skew rate to be nearly zero. Additionally, considering the generated image's dimensions of $512^2$, we assigned $c_x=c_y=256$. The most important value was the focal length, that is, $f_x$ and $f_y$. As these values are typically quite similar, we assumed equivalence and conducted multiple investigations to verify this assumption.

\begin{figure*}
\begin{center}
\includegraphics[width=1\linewidth]{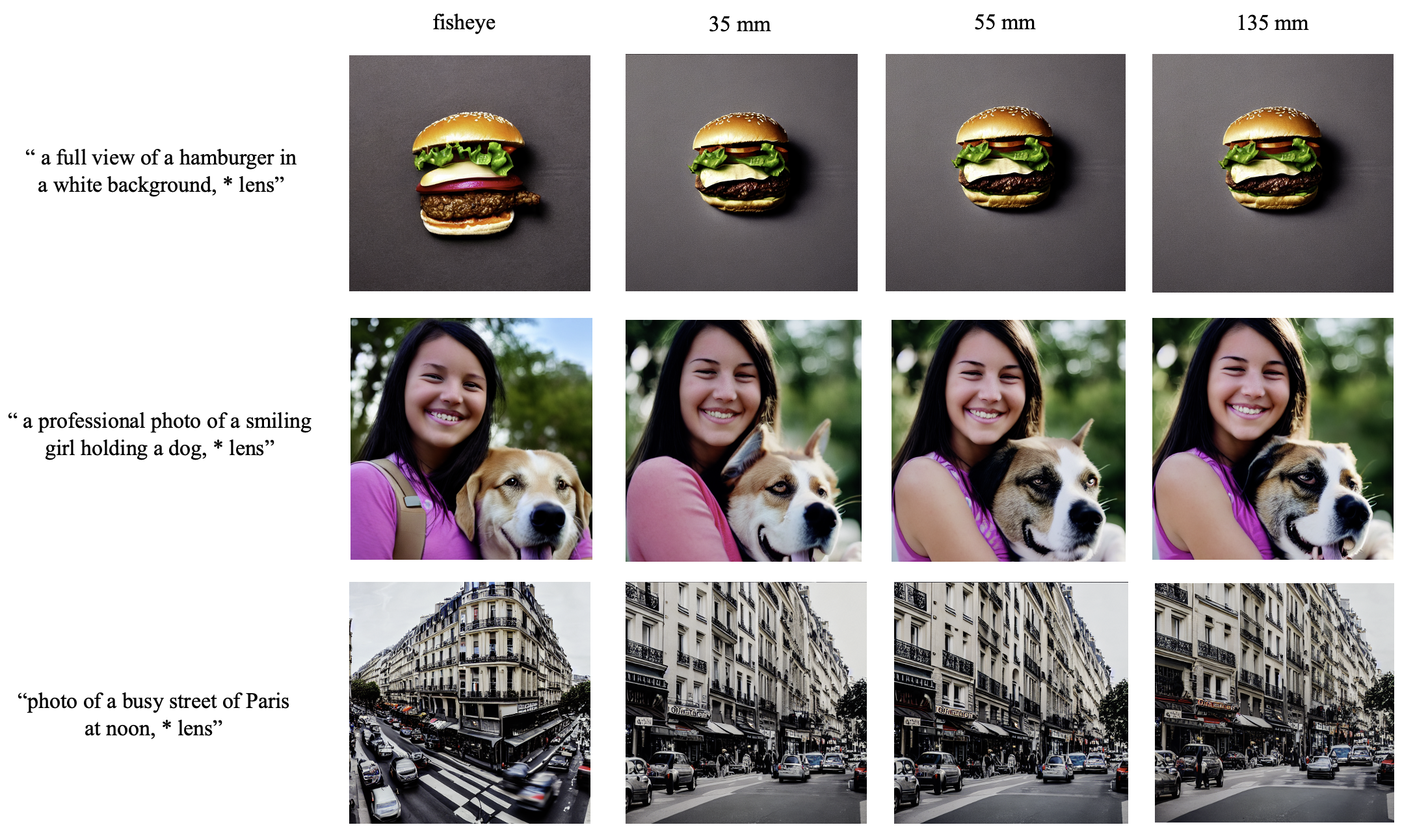}
\end{center}\vspace{-0.6cm}
   \caption{Focal length test using Stable diffusion~\cite{rombach2022high}. We observed differences between generated images with additional lens information. There are no substantial differences in the result.}
\label{fig:short}
\label{focal}
\end{figure*}

Fig.~\ref{focal} illustrates the incorporation of an additional prompt for lens focal length alongside the three existing prompts: object, portrait, and landscape. The focal length prompts employed in this instance include "fish eye," "35mm lens," "50mm lens," and "135mm lens". Upon examining the outcomes, we observed no substantial differences in the results, except the landscape category, particularly the "fish eye" prompt. This phenomenon is attributed to the majority of training data falling within the range of 35mm to 50mm focal lengths. Consequently, we generated point clouds assuming a focal length of approximately 45mm, yielding satisfactory results for the majority of tasks necessitating the generation of object-related images.

\subsection{Progressive global view sampling (PGVS)}
The beta distribution for PGVS is initialized with $\alpha_0=2$ and $\beta_0=8$. Thereafter, both constants decrease linearly until the $t_u$ iteration, which is defined as $0.3$ times the total number of iterations $t_{total}$. After $t_u$, both constants are fixed at a value of 1 for the remaining iterations. For the experiments reported in this paper, we set $t_{total}$ to 5000, resulting in $t_u$ being equal to 1500.

\subsection{Sparsity loss}
The sparsity loss used in this study was introduced in the Dreamfields~\cite{jain2021dreamfields} and was also employed in Latent-NeRF. It has a similar form to the binary cross-entropy loss and is designed to encourage the density values to converge to either 0 or 1. This helps improve the quality of the generated 3D representation by ensuring that the rays either pass through completely or are blocked. The mathematical expression for the sparsity loss is shown below.

\small
\begin{align}
\mathcal{L}_{sp} = - \mathbb{E}[\alpha log(\alpha) + (1-\alpha)log(1-\alpha)]
\end{align}
\normalsize
with $\alpha$ is the weighted sum of a ray which is clipped with $[\epsilon, 1-\epsilon]$, where $\epsilon = 10^{-5}$ and $\lambda_{sp} = 5\cdot10^{-5}$.

\subsection{Reliability-guided loss}
The equation for reliability-guided loss is on the following:
\small
\begin{align}
\mathcal{L}_R = [\zeta \cdot \mathcal{M} + \eta(t) \cdot (1 - \mathcal{M})] \odot || z_{r} - z_{p} ||_{1}
\end{align}
\normalsize
with $\eta(t) = e^{-t/\lambda_\eta}$. And all the experiments proceeded with $\lambda_{\eta}=8$, $\zeta=2$.

\subsection{Patch refinement}
Patch refinement is a crucial method to ensure that the final 3D representation has uniformity in terms of overall tone and texture. This refinement applies $N_{patch}$ patches of size $k \times k$ to a mask with size $H \times W$. The choice of patch size $k$ is important. In the case of large $k$, it shows the unintended result such as a change of a semantic part of the existing IB. When $k$ goes too small, the effect of patch refinement would disappear while downsizing the mask. Similarly, the number of patches $N_{patch}$ is also important, as an insufficient number of patches may not produce the appropriate refinement, while an excessive number of patches may not maintain the existing IB. Therefore, we selected the values of $k = 16$ and $N = 256$ for this experiment.

\section{Implementation Details}
\subsection{Monocular depth estimation}
The image/text-to-3D task at hand necessitates the ability to extract accurate relative depth maps from a set of images that consists of diverse objects. Hence, it was crucial to employ a robust and accurate depth estimation model that had been trained on a variety of datasets, and for this purpose, we adopted MiDaS~\cite{Ranftl2022}. MiDaS was trained using pre-existing models from 12 different datasets, including ReDWeb~\cite{Xian_2018_CVPR}, DIML~\cite{kim2018deep}, Movies, MegaDepth~\cite{MegaDepthLi18}, WSVD~\cite{wang2019web}, TartanAir~\cite{tartanair2020iros}, HRWSI~\cite{Xian_2020_CVPR}, ApolloScape~\cite{huang2019apolloscape}, BlendedMVS~\cite{yao2020blendedmvs}, IRS~\cite{wang2019irs}, KITTI~\cite{Geiger2013IJRR}, and NYU Depth V2~\cite{Silberman:ECCV12}. The largest model provided by MiDaS `DPT BeiT Large'~\cite{Ranftl2021} which offers the highest quality depth estimation among the available models, was employed. This model utilizes a transformer architecture, which enables more precise and detailed depth estimation as compared to convolutional structures.

\subsection{IB 3D construction based on depth map}
\begin{figure}[!htb]
    \centering
    \vspace{-0.5em}
    \includegraphics[width=0.9\linewidth]{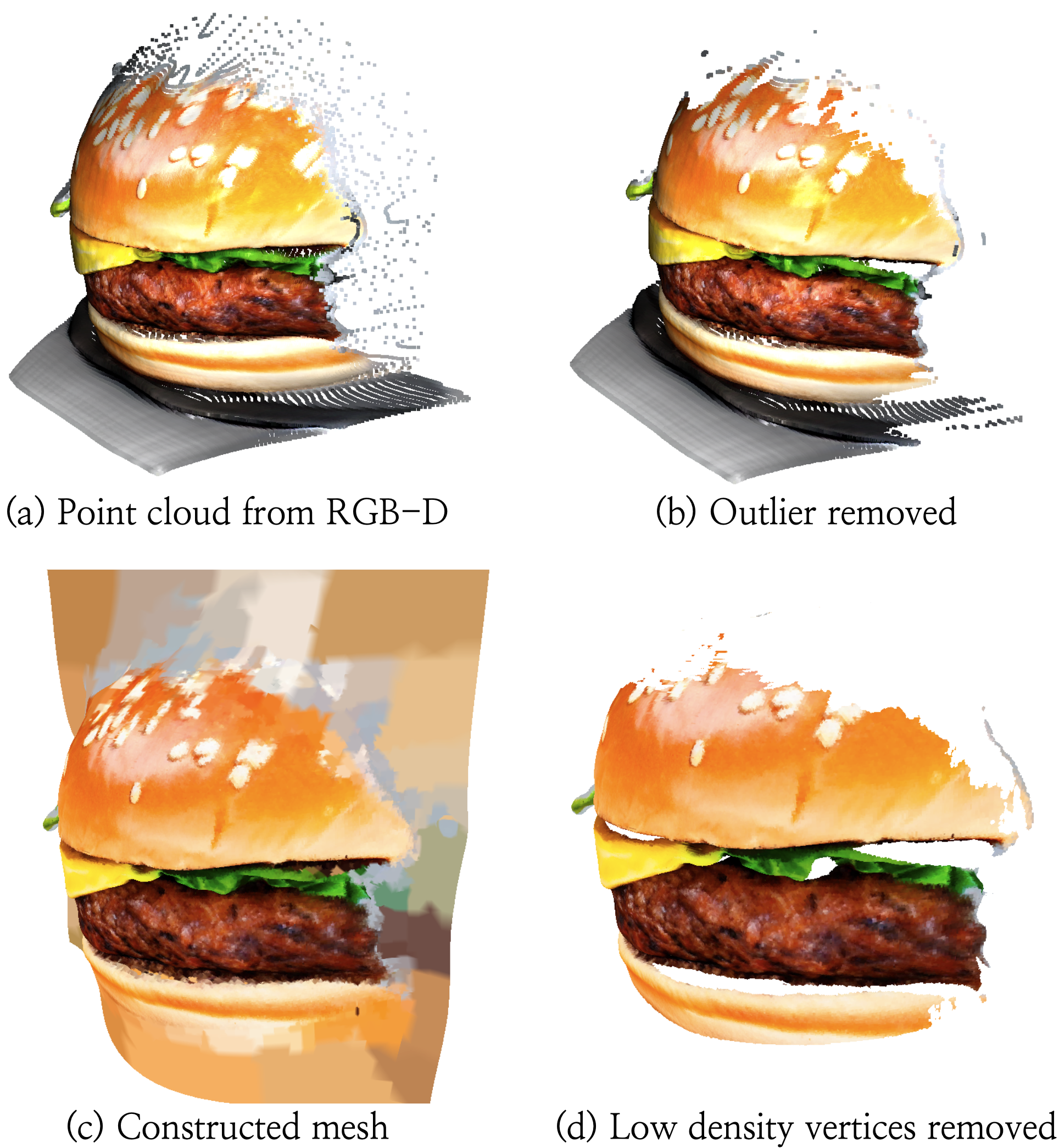}
    \caption{Outputs of each step in IB 3D construction. An example of established point cloud from RGB-D image is in (a). Representation of point cloud after removing outliers is shown in (b). By Poisson reconstruction, mesh (c) is created. And lastly, after removing low density vertices, we get final result like (d).}
    \label{fig_IB3D}
\end{figure}

An RGB-D image can be generated by combining a depth map obtained from monocular depth estimation with an image, followed by the creation of a point cloud using the Open3D~\cite{zhou2018open3d} library. To eliminate outliers, points are removed if their distance from the five surrounding points exceeds one standard deviation. The normal vector of each vertex is then estimated from the resulting point cloud, and a mesh is created through Poisson surface reconstruction, with a depth value of 10. Vertices with a density below 0.1 quantiles are subsequently removed, based on the assumption that the object of interest is expected to exhibit a relatively high density. An example of the process's output is shown in Fig.~\ref{fig_IB3D}

\subsection{Stable diffusion inpainting}
We utilized Stable diffusion inpainting~\cite{rombach2022high} for inpainting-LDM to train NeRF in OB. The model was fine-tuned with original Stable diffusion. The following steps shows how it was fine-tuned:
To initialize the Stable diffusion inpainting model, the creater used the weights from the Stable diffusion-v-1-2 model. The training process consisted of two phases: regular training for 595k steps followed by 440k steps of training with inpainting task at a resolution of 512 $\times$ 512 using the `LAION-Aesthetics v2 5+' dataset~\cite{schuhmann2022laion}. To improve the classifier-free guidance sampling, the text-conditioning was dropped by 10\%. For inpainting, the UNet was augmented with five additional input channels - four for the encoded masked-image and one for the mask itself. These additional channels were zero-initialized after restoring the non-inpainting checkpoint. During the training process, synthetic masks were generated and 25\% of the images were masked in each iteration.

\subsection{Prompt conditioning}
Monocular depth estimation is typically carried out by leveraging the shadow information present in an image, along with prior knowledge incorporated into the model. As a consequence, the depth map generated by such methods may be inadequate for images that lack substantial shading. Furthermore, it has been observed that the depth estimation process encounters difficulties when the background is either complex or positioned in close proximity to the subject. To overcome these issues, we propose a preprocessing step in which the initial image used for generating an IB 3D object is accompanied by the phrase "A whole photo of $\sim$" at the beginning of the prompt for non-cropped image and "$\sim$ in the white background taken with 50mm lens" at the end of the prompt. This ensures that the depth estimation process proceeds smoothly, resulting in an acceptable IB 3D object.

\section{Experimental Details}

\subsection{Evaluation details}
\paragraph{Baselines.}
In this study, we conducted a comparative analysis of existing image-to-3D models and text-to-3D models against our proposed DITTO-NeRF model. For the image-to-3D model, we evaluated our results against the current state-of-the-art model, NeuralLift-360~\cite{Xu_2022_neuralLift}, for which the code has not been made publicly available. Therefore, we compared our results with the reference images provided on the NeuralLift-360 webpage~\cite{xu2023Neuralliftweb}. Since the size of the NeuralLift-360 results was small, we cropped them to ensure a fair quality comparison. Regarding the text-to-3D model, we compared our results with two existing models based on Stable diffusion: Stable-Dreamfusion~\cite{stable-dreamfusion} and Latent-NeRF~\cite{metzer2022latent}. Stable-Dreamfusion replaces the generative diffusion model with Stable diffusion instead of Imagen which was used for the original Dreamfusion~\cite{poole2022dreamfusion}. In contrast, Latent-NeRF learns the 3D representation directly in the latent space by learning the gradient through Stable diffusion in latent space. During the evaluation, the rendered latent vector is passed through the decoder of a Variational AutoEncoder (VAE) to extract the final RGB image, thereby achieving a fast learning time. All the evaluation of Stable-dreamfusion was conducted on $20000$ iterations because the $10000$ iterations' output quality was not enough for proper comparison and Latent-NeRF, it was $5000$ iterations. Our work was conducted based on the code provided by Latent-NeRF. 

\paragraph{User study.}
In this study, a total of 3,150 responses were collected through administering a survey of 15 questions to 210 participants. The Image-to-3D evaluation was performed using four different models, namely, SinNeRF~\cite{xu2022sinnerf}, DietNeRF~\cite{jain2021putting}, NeuralLift-360~\cite{Xu_2022_neuralLift}, and our DITTO-NeRF about three different images, namely, "baseball" "apple" and "hydrant" all of which were sourced from the NeuralLift-360 webpage. The evaluation of the image-to-3D object correspondence and fidelity was carried out with respect to the models' outcome and scored on a five-point scale. Additionally, the text-to-3D task was evaluated for diversity using three prompts, namely, "a loaf of bread", "a small saguaro cactus planted on a clay pot" and "a single candle burning on an ornate silver candlestick". Outputs were generated by Stable-Dreamfusion, Latent-NeRF, and our DITTO-NeRF with three different seeds for each of the three prompts. The evaluation procedure was similar to that used in the image-to-3D evaluation and the diversity of the outputs was scored on a five-point scale. Lastly, the fidelity of the outputs in the text-to-3D task was evaluated using three different prompts, namely, "a hamburger" "an astronaut" and "a suitcase" with the same three models. The results of each evaluation item were averaged to create mean opinion scores.

\section{Additional Ablation Studies}
\subsection{Effectiveness of Dimension refinement}
\begin{figure}[!htb]
    \centering
    \includegraphics[width=1.\linewidth]{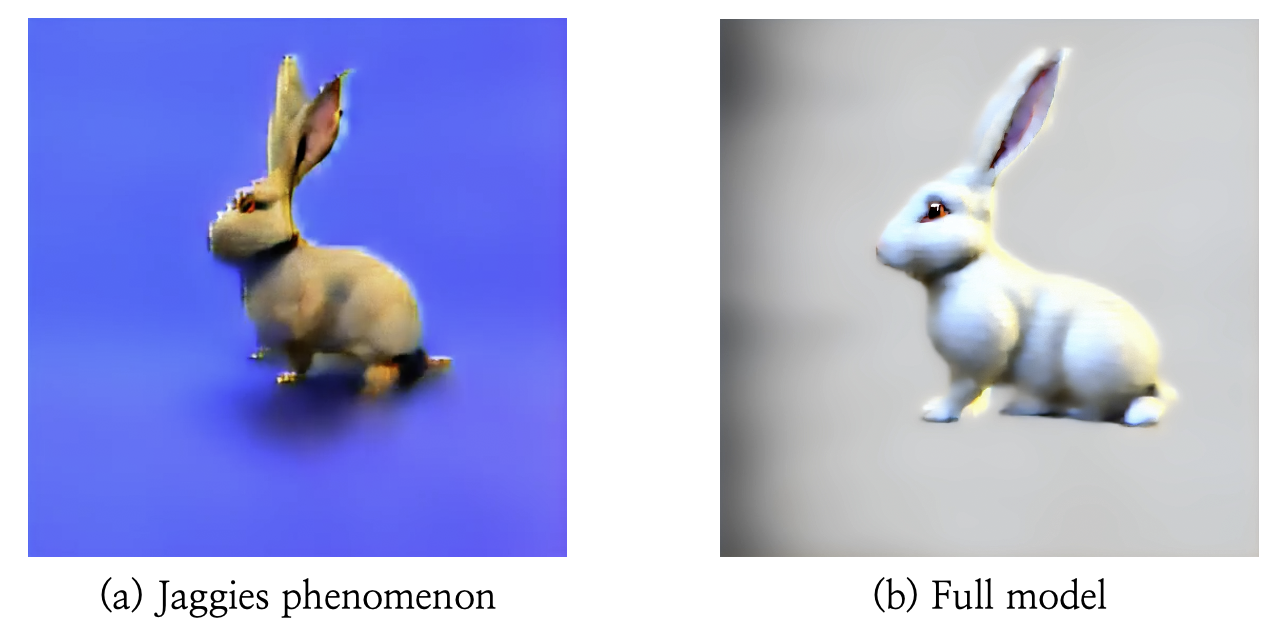}
    \vspace{-1em}
    \caption{An example of jaggies problem is shown in (a). With Dimension refinement, we can see there is no jaggies problem in (b).}
    \vspace{-1em}
    \label{fig_supp_jagg}
\end{figure}

If the dimension refinement stage does not exist, we can see ‘Jaggies’ phenomenon where the boundary looks like a saw, as shown in Fig.~\ref{fig_supp_jagg}. As the cause of this, we identified that it is due to the projection of high-frequency information such as pre-rendered images into a low dimension such as latent space. We devised a method to increase the rendering dimension of NeRF to alleviate these artifacts. However, there was a problem that the training time increased rapidly when continuously using a large rendering dimension. Therefore, we devised a method of linearly increasing the rendering dimension during the refinement stage, finally doubling it, focusing on the fact that coarse shapes are generated even at low rendering dimensions. In actual implementation, the rendering dimension starting at $64$ starts to increase linearly in the refinement stage and finally reaches $128$ and learning ends.

\subsection{Effectiveness of PGVS}
We tried other types of distributions for view sampling such as simple uniform sampling, `moving beta distribution' and `discrete accumulation distribution'. In this chapter, we would like to explain the two distributions mentioned above and their results.

\begin{figure}[!htb]
    \centering
    \includegraphics[width=1.\linewidth]{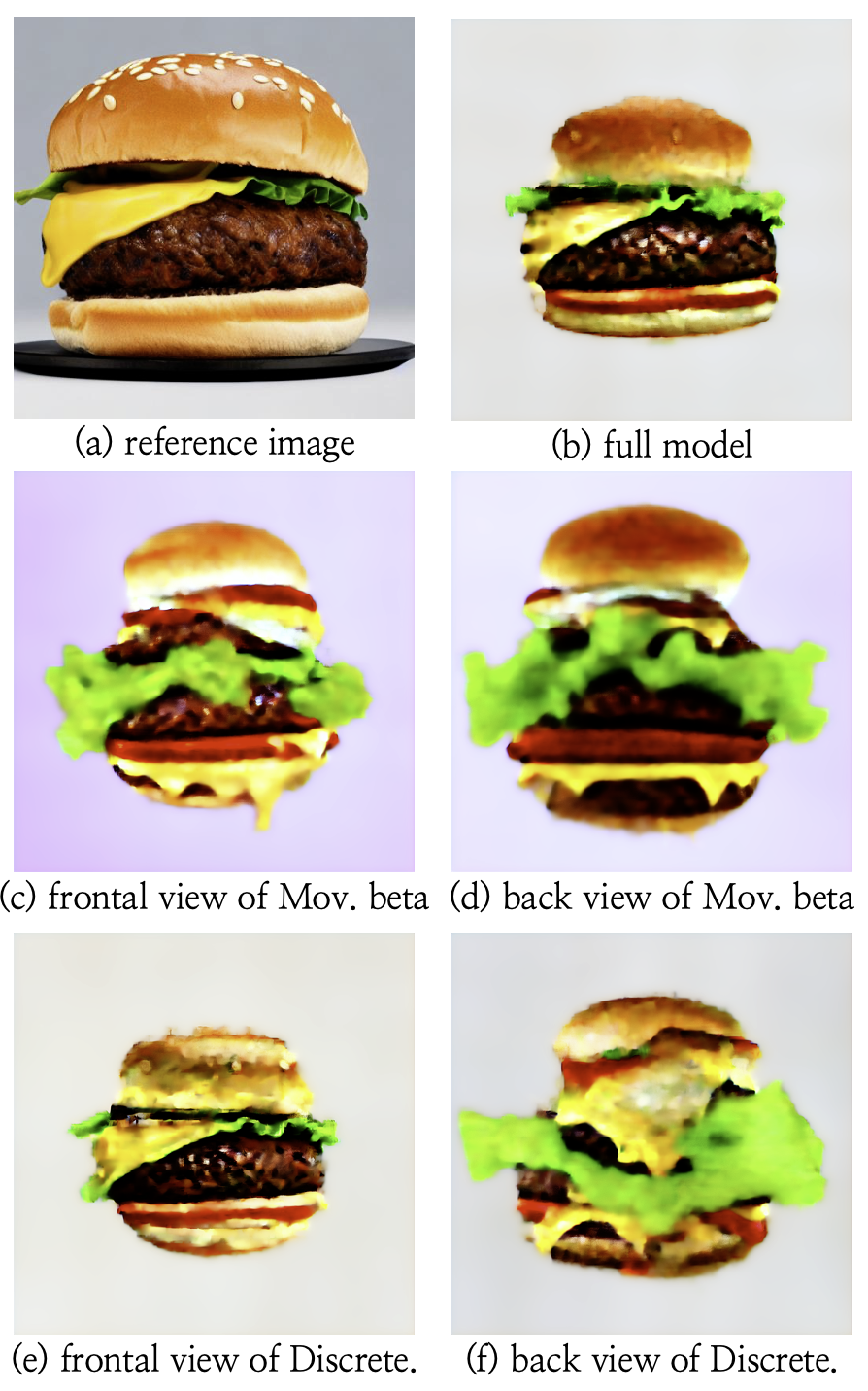}
    \vspace{-1em}
    \caption{Comparison of results with different camera view sampler. (a) is a reference image generated by latent diffusion model with a prompt "a hamburger". (b) is the result of our full model, DITTO-NeRF. (c) is the frontal view of the result with Moving beta distribution sampling. (d) shows back view of the Moving beta distribution sampling. And (e) shows the frontal view of the result when Discrete accumulation sampling was utilized. (f) refers to the back view of the results with Discrete accumulation view sampling.}
    \vspace{-1em}
    \label{fig_supp_sampler}
\end{figure}

\paragraph{Moving beta distribution.} In our full model, the Probabilistic Gradient Vector Sampling (PGVS) method is employed to optimize the camera placement. In this method, the values of $\alpha$ and $\beta$ of the beta distribution decrease simultaneously and eventually converge to a value of 1. However, in the modified distribution used in this study, the two values gradually change to each other's first value. During the initial phase of training, many camera positions are sampled in the frontal part, and as the training progresses, a larger number of samples are taken in the back part. Similar to PGVS, the modified distribution also has a uniform point, after which the camera view is sampled uniformly. The results of this approach are presented in Fig.~\ref{fig_supp_sampler}, where it can be observed that the frontal reconstruction did not work as effectively as in the full model.

\paragraph{Discrete accumulation distribution.} The sampler used in this study is similar to the moving beta distribution described earlier, but it differs in that it has a discrete distribution. This sampler operates within a pre-defined interval, where the probability within a specific interval is set to $r$, and the probability in the remaining intervals is set to $1-r$. At each specific iteration, the interval with the probability of $r$ is passed on to the next iteration. Although the IB part was reconstructed effectively using this sampler, the back part did not converge to a normal shape. To address this issue, an experiment was conducted where $r$ was set to 0.65, as shown in Fig.~\ref{fig_supp_sampler}.

{\small
\bibliographystyle{ieee_fullname}
\bibliography{egbib}
}

\end{document}